\documentclass{article}
\usepackage{amsmath,amssymb,amsbsy,amsthm}

\let\epsilon\varepsilon
\let\phi\varphi

\def\X{{\cal X}}
\def\N{\mathbb N}

\def\C{\mathcal C}
\def\E{{\bf E}}

\def\-as{\text{-a.s.}}
\def\argmax{\operatorname{argmax}}

\newtheorem{theorem}{Theorem}
\newtheorem{definition}{Definition}

\begin{document}
\title{Characterizing predictable classes of processes}
\author{{\bf Daniil Ryabko} \\  INRIA Lille-Nord Europe,
\\   daniil@ryabko.net
}
\maketitle

\begin{abstract} 
The problem is sequence prediction in the following setting. 
A sequence $x_1,\dots,x_n,\dots$ of discrete-valued observations is generated 
according to some unknown probabilistic law (measure) $\mu$. After observing each outcome, 
it is required to give the conditional probabilities of the next observation.
The measure  $\mu$ belongs to an arbitrary class $\C$  of stochastic processes.
We are interested in predictors $\rho$ whose conditional probabilities converge to the
``true'' $\mu$-conditional probabilities if any $\mu\in\C$ is chosen to generate the data.
We show that if such a predictor exists, then a predictor can also be obtained as a convex
combination of a countably many elements of $\C$. In other words, it can be obtained as a 
Bayesian predictor whose prior is concentrated on a countable set.  This result is established for two very different
measures of performance of prediction, one of which is very strong, namely, total variation, and the other is very weak, 
namely, prediction in expected average  Kullback-Leibler divergence.

\end{abstract}
\section{Introduction}
Given a  sequence $x_1,\dots,x_n$ of observations $x_i\in\X$, where $\X$ is a finite set, we 
want to predict what are the probabilities of observing $x_{n+1}=x$ for each $x\in\X$, before $x_{n+1}$ is revealed,
after which the process continues.
It is assumed that the sequence is generated by some unknown stochastic process $\mu$, a probability measure
on the set of one-way infinite sequences $\X^\infty$. The goal is to have a predictor whose predicted probabilities
converge (in a certain sense) to the correct ones (that is, to $\mu$-conditional probabilities). In general this goal is impossible to achieve if 
nothing is known about the measure $\mu$ generating the sequence. In other words, one cannot have a predictor
whose error goes to zero for any measure $\mu$. The problem becomes tractable if we assume that the measure $\mu$
generating the data belongs to some known class $\C$.
The  questions addressed in this work are a part of the following general problem: given an arbitrary set  $\C$ of measures, how can we find 
a predictor that performs well when the data is generated by any  $\mu\in\C$, and whether
it is possible to find such a predictor at all. 
An example of a generic  property  of a class $\C$ that allows for construction of a predictor, is that
$\C$ is countable. Clearly, this condition is very strong. An example,  important from the applications point of view,  of a class $\C$ of measures  
for which  predictors are known,  is the class of all stationary measures. The general question, however, is very far from being answered.

The contribution of this work to solving this question is in that we 
we provide a specific form in which to look for a solution to  the general problem.
More precisely, we show that if a predictor exists, 
then a predictor  can also be obtained as a weighted sum of a countably many elements of $\C$.  This result
can also be viewed as a justification of the Bayesian approach to sequence prediction: if there exists 
a predictor which predicts well every measure in the class, then there exists a Bayesian predictor (with a rather simple prior) that has this property too.
In this respect it is important to note that the result obtained about such a  Bayesian predictor
is pointwise (holds for every $\mu$ in $\C$), and stretches far beyond the set its prior is concentrated on.

The {\bf motivation} for studying predictors for arbitrary classes $\C$ of processes is two-fold. First of all, prediction is a basic ingredient for 
constructing intelligent systems. Indeed, in order to be able to find optimal behaviour in an unknown environment,
an intelligent agent must be able, at the very least, to predict how the environment is going to behave (or, to be more precise, how relevant 
parts of the environment are going to behave).
Since the response of the environment may in general depend on the actions of the agent, this response
is necessarily non-stationary for explorative agents. Therefore, one cannot readily use prediction methods developed for stationary 
environments, but rather has to find predictors for the classes of processes that can appear as a possible response
of the environment. 

Apart from this, the problem of prediction itself has numerous applications in such diverse
fields as data compression, market analysis,  bioninformatics, and many others. It seems clear that prediction methods constructed
for one application cannot be expected to be optimal when applied to another. Therefore, an important question is how to develop
specific prediction algorithms for each of the domains. In order to do this, the first step is to understand 
for which classes of problems (i.e. sets of measures generating the data) a predictor exists.

{\bf Prior work}.
As it was mentioned,  if the class $\C$ of measures is countable (that is, if $\C$ can be represented as $\C:=\{\mu_k: k\in\N\}$), then 
there exists a predictor which performs well for any $\mu\in\C$. Such a predictor can be obtained as a  Bayesian mixture 
$\rho_S:=\sum_{k\in\N} w_k \mu_k$, where $w_k$ are summable positive real weights, and it has very strong predictive properties; in particular,  $\rho_S$ predicts every $\mu\in\C$ in total variation distance, as follows from the result of  \cite{Blackwell:62}.
Total variation distance measures the difference in (predicted and true) conditional probabilities of all future 
events, that is, not only the probabilities of the next observations, but also of observations that are arbitrary far off in the future (see formal 
definitions below).  In the context of sequence prediction the measure $\rho_S$  was first studied 
by  \cite{Solomonoff:78}. Since then, the idea of taking a convex combination of a finite or countable class of measures (or predictors) to
obtain a predictor permeates most of the research on sequential prediction (see, for example, \cite{Cesa:06}) and some related topics in AI~\cite{Hutter:04uaibook, Ryabko:08ao++}. 
In practice it is clear that, on the one hand, countable models are not sufficient, since already the class $\mu_p, p\in[0,1]$ of
Bernoulli i.i.d. processes, where $p$ is the probability of 0, is not countable. On the other hand, prediction in total variation
can be too strong to require; predicting probabilities of the next observation may be sufficient, maybe even not 
on every step but in the Cesaro sense. A key observation here is that a predictor $\rho_S=\sum w_k\mu_k$ may be a good predictor
not only when the data is generated by one of the processes $\mu_k$, $k\in\N$, but when it comes from a much larger class.
Let us consider this point in more detail. Fix for simplicity $\X=\{0,1\}$. The Laplace predictor
$
 \lambda(x_{n+1}=0|x_1,\dots,x_n)=\frac{\#\{i\le n: x_i=0\}+1}{n+|\X|}
$ 
predicts any Bernoulli i.i.d.~process: although convergence in total variation distance of conditional probabilities
does not hold, predicted probabilities of the next outcome converge to the correct ones.
 Moreover, generalizing the Laplace predictor,  a predictor  $\lambda_k$ can be constructed for 
the class $M_k$ of all $k$-order Markov measures, for any given $k$. As was found by  \cite{BRyabko:88}, the combination $\rho_R:=\sum w_k\lambda_k$
is a good predictor not only for the the set $\cup_{k\in\N} M_k$ of all finite-memory processes, but also for any measure
$\mu$ coming from a much larger class: that of all stationary measures on $\X^\infty$. Here prediction is possible
only in the Cesaro sense (more precisely, $\rho_R$ predicts every stationary process in expected time-average  Kullback-Leibler divergence,
see definitions below).
The Laplace predictor itself can be obtained as a Bayes mixture over all Bernoulli i.i.d. measures with uniform 
prior on the parameter $p$ (the probability of 0). However, as was observed in \cite{Hutter:07upb} (and as is easy to see),
the same (asymptotic) predictive properties are possessed by  a Bayes mixture with a countably supported prior which is dense in 
$[0,1]$ (e.g. taking $\rho:=\sum w_k \delta_k$ where $\delta_k, k\in\N$ ranges over all Bernoulli i.i.d. measures with rational probability of 0). 
For a given $k$, the set of $k$-order Markov processes is parametrized by finitely many $[0,1]$-valued parameters. Taking a dense 
subset of the values of these parameters, and a mixture of the corresponding measures, results in a predictor for the class
of $k$-order Markov processes. Mixing over these (for all $k\in\N$) yields, as in \cite{BRyabko:88}, a predictor for the class of all 
stationary processes. Thus, for the mentioned classes of processes, a predictor can be obtained as a Bayes mixture of 
 countably many measures in the class. An additional reason why this kind  of analysis is interesting is because of the difficulties arising in trying to construct 
Bayesian predictors for classes of processes that can not be easily parametrized. Indeed, a natural way to obtain 
a predictor for a class $\C$ of stochastic processes is to take a Bayesian mixture of the class. 
To do this, one needs to define the structure of a probability space on $\C$. 
If the class $\C$ is well parametrized, as is the case with the set of all Bernoulli i.i.d. process, then one can 
integrate with respect to the parametrization. In general, when the problem lacks a natural parametrization, although one can define the structure of the probability 
space on the set of (all) stochastic processes in many different ways, the results one can obtain will then be
with probability 1 with respect to the prior distribution (see, for example, \cite{Jackson:99}), while pointwise consistency
cannot be assured (see e.g. \cite{Diaconis:86}). 
Results with prior probability 1  can be hard to interpret if one is not sure that the structure 
of the probability space defined on the set $\C$ is indeed a natural one for the problem at hand (whereas if one does have a natural parametrization,
then usually results for every value of the parameter can be obtained, as in the case with Bernoulli i.i.d. processes mentioned above).
The results of the present work show that when a predictor exists it can indeed be given as  a Bayesian 
predictor, which predicts  every (and not almost every) measure in the class, while its support is only countable.

{\bf The results.} Here we show that if there is a predictor
that performs well for every measure coming from a class $\C$ of processes, then a predictor can also be obtained as a convex combination $\sum_{k\in\N} w_k\mu_k$ for some $\mu_k\in\C$ and
some $w_k>0$, $k\in\N$. 
This holds if the prediction quality is measured by either total variation distance, or expected average KL divergence: 
one measure of performance that is very strong, the other rather weak. 
The analysis for the total variation case 
relies on the fact that if $\rho$ predicts $\mu$ in total variation distance, then $\mu$ is absolutely continuous
with respect to $\rho$, so that $\rho(x_{1..n})/\mu(x_{1..n})$ converges to a positive number with $\mu$-probability 1
and with a positive $\rho$-probability. However, if we settle for a weaker measure of performance, such as  expected average KL divergence, measures $\mu\in\C$ are typically singular with 
respect to a predictor $\rho$. Nevertheless, since $\rho$ predicts $\mu$ we can show that  $\rho(x_{1..n})/\mu(x_{1..n})$ decreases subexponentially
with $n$ (with hight probability), and then we can use this ratio as an analogue of the density for each time step $n$, and 
find a convex combination of countably many measures from $\C$ that has  desired
predictive properties for each $n$. Combining these predictors for all $n$ then results in a predictor that predicts every $\mu\in\C$ in average KL divergence.  The proof techniques developed  have a potential
to be used in solving other questions concerning sequence prediction, in particular, the general question of how to find a predictor
for an arbitrary class $\C$ of measures.

\section{Preliminaries}
Let $\X$ be a finite set. The notation $x_{1..n}$ is used for $x_1,\dots,x_n$. 
 We consider  stochastic processes (probability measures) on $(\X^\infty,\mathcal F)$ where $\mathcal F$
is the sigma-field generated by the cylinder sets  $[x_{1..n}]$, $x_i\in\X, n\in\N$, 
where $[x_{1..n}]$ is the set of all infinite sequences that start with $x_{1..n}$.
For a  finite set $A$ denote $|A|$ its cardinality.
We use  $\E_\mu$ for
expectation with respect to a measure $\mu$.

Next we introduce the measures of the quality of prediction used in this paper.
For two measures  $\mu$ and $\rho$  we are interested in how different 
 the $\mu$- and $\rho$-conditional probabilities are, given a data sample $x_{1..n}$.
Introduce the {\em total variation} distance 
$$
v(\mu,\rho,x_{1..n}):= \sup_{A\in\mathcal F} |\rho(A|x_{1..n})-\mu(A|x_{1..n})|.
$$
\begin{definition}
We say that $\rho$ predicts $\mu$ in total variation if 
$$
v(\mu,\rho,x_{1..n})\to0\ \mu\-as
$$
\end{definition}
This convergence is rather strong. In particular, it means that $\rho$-conditional probabilities
of arbitrary far-off events converge to $\mu$-conditional probabilities. 
Moreover,  $\rho$ predicts $\mu$ in
total variation if \cite{Blackwell:62} and only if \cite{Kalai:94} $\mu$ is absolutely continuous with respect to $\rho$.

Thus, for a class $\C$ of measures there is a predictor $\rho$ that predicts every $\mu\in\C$ in total
variation if and only if every $\mu\in\C$ has a density with respect to $\rho$.
Although such  sets of processes are rather large, they do not include even such basic 
examples as the set of all Bernoulli i.i.d. processes.
That is, there is no $\rho$ that would predict in total variation every Bernoulli i.i.d. process measure $\delta_p$, $p\in[0,1]$,
where $p$ is the probability of $0$. 
Therefore, perhaps for many (if not most) practical applications this measure of the quality of prediction is too strong,
and one is interested in weaker measures of performance.

For two measures $\mu$ and $\rho$ introduce the {\em expected cumulative Kullback-Leibler divergence (KL divergence)} as
\begin{equation}\label{eq:akl} 
  d_n(\mu,\rho):=  \E_\mu
  \sum_{t=1}^n  \sum_{a\in\X} \mu(x_{t}=a|x_{1..t-1}) \log \frac{\mu(x_{t}=a|x_{1..t-1})}{\rho(x_{t}=a|x_{1..t-1})},
\end{equation}
In words, we take the expected (over data) average (over time) KL divergence between $\mu$- and $\rho$-conditional (on the past data) 
probability distributions of the next outcome.
\begin{definition}
We say that $\rho$ predicts $\mu$ in expected average KL divergence if 
$$
{1\over n} d_n(\mu,\rho)\to0.
$$
\end{definition}
This measure of performance is much weaker, in the sense that it requires good predictions only one step ahead, and not on every step
but only on average; also the convergence is not with probability 1 but in expectation. With prediction quality so measured, 
predictors  exist for relatively large
classes of measures; most notably, \cite{BRyabko:88} provides a predictor which predicts every stationary 
process in expected average KL divergence. 
A simple but useful identity that we will need (in the context of sequence prediction introduced also in \cite{BRyabko:88})
is the following
\begin{equation}\label{eq:kl}
 d_n(\mu,\rho)=-\sum_{x_{1..n}\in\X^n}\mu(x_{1..n}) \log \frac{\rho(x_{1..n})}{\mu(x_{1..n})},
\end{equation}
where on the right-hand side we have simply the KL divergence between measures $\mu$ and $\rho$ restricted to the first $n$ observations.

Thus, the results of this work will be established with respect to two very different measures of prediction quality,
one of which is very strong and the other rather weak. This suggests that the facts established reflect some fundamental
properties of the problem of prediction, rather than those pertinent to particular measures of performance. On the other hand,
it remains open to extend the results below to different measures of performance.

\section{Main results}

\begin{theorem}\label{th:1} Let $\C$ be a set of probability measures on $\X^\infty$. If there is a measure $\rho$ such that $\rho$ predicts every $\mu\in\C$ in 
total variation, then there is a sequence $\mu_k\in\C$, $k\in\N$ such that the measure $\nu:=\sum_{k\in\N} w_k\mu_k$ predicts every $\mu\in\C$ in 
total variation, where $w_k$ are any positive weights  that sum to 1.
\end{theorem}
This relatively simple fact can be proven in different ways, relying on the equivalence 
of the statements ``$\rho$ predicts $\mu$ in total variation distance'' and ``$\mu$ is absolutely continuous
with respect to $\rho$.'' The proof presented below uses techniques that can be then generalized to 
the case of  prediction in expected average KL-divergence, where in all interesting cases 
all measures $\mu\in\C$ are singular with respect to any predictor that predicts all of them.
The idea of the proof of Theorem~\ref{th:1} is as follows. For each measure $\mu\in\C$ we
find the set $T_\mu$  of sequences $x_1,x_2,\dots$ on which the density of $\mu$ with respect to $\rho$ exists and is non-zero.
Such a set has $\mu$-probability 1, and,  by absolute continuity, a positive $\rho$-probability. The idea is then to cover the union $\cup_{\mu\in\C}T_\mu$ with countably
many of these sets, and then construct a new predictor as a sum of the corresponding measures. To find this countable
collection of sets $T_\mu$, we first find a largest (up to an  $\epsilon_1$) one with respect $\rho$, then the one who has a largest 
(up to an  $\epsilon_2$)
part not covered by the first set, and so on (where $\epsilon_k$ are decreasing). Then we show that any strictly convex combination of the resulting sequence of measures
has the property that any measure in $\C$ is absolutely continuous with respect to it.
\begin{proof}
We break the (relatively easy) proof of this theorem into 3 steps, which will make the (more involved) proof
of the next theorem more understandable.

\noindent{\em Step 1: densities.}
 For any $\mu\in\C$, since $\rho$ predicts $\mu$ in total variation, $\mu$ has a density (Radon-Nikodym derivative) $f_\mu$ with respect 
to $\rho$. Thus, for the set $T_\mu$ of all sequences $x_1,x_2,...\in\X^\infty$ on which $f_\mu(x_{1,2,\dots})>0$ 
(the limit $\lim_{n\rightarrow\infty}\frac {\rho(x_{1..n})}{\mu(x_{1..n})}$ 
exists and is finite and positive) we have $\mu(T_\mu)=1$ and $\rho(T_\mu)>0$. Next we will construct a sequence of measures $\mu_k\in\C$, $k\in\N$ such that 
the union of the sets $T_{\mu_k}$ has probability 1 with respect to every $\mu\in\C$, and will show that this is a sequence of measures whose existence is asserted in the theorem statement.

{\em Step 2: a countable cover and the resulting predictor.}
Let $\epsilon_k:=2^{-k}$ and let $m_1:=\sup_{\mu\in\C}\rho(T_\mu)$. Clearly, $m_1>0$. Find any $\mu_1\in\C$ such that $\rho(T_{\mu_1})\ge m_1-\epsilon_1$, and let
$T_1=T_{\mu_1}$. For $k>1$ define $m_k:=\sup_{\mu\in\C}\rho(T_\mu\backslash T_{k-1})$. If $m_k=0$ then define $T_{k}:=T_{k-1}$, otherwise find any $\mu_k$ such 
that $\rho(T_{\mu_k}\backslash T_{k-1})\ge m_k-\epsilon_k$, and let $T_k:=T_{k-1}\cup T_{\mu_k}$. 
Define the predictor $\nu$ as $\nu:=\sum_{k\in\N}w_k\mu_k$.

{\em Step 3: $\nu$ predicts every $\mu\in\C$.}
Since the sets 
$T_1$, $T_2\backslash T_1,\dots, T_k\backslash T_{k-1},\dots$ are disjoint, 
we must have $\rho(T_{k}\backslash T_{k-1})\to0$, so that $m_k\to0$.
Let 
$$
T:=\cup_{k\in\N} T_k.
$$ 
Fix any $\mu\in\C$. 
Suppose that  $\mu(T_{\mu} \backslash T)>0$. Since $\mu$ is absolutely continuous
with respect to $\rho$, we must have $\delta:=\rho(T_{\mu}\backslash T)>0$. Then for every $k>1$ we have
$$m_k=\sup_{\mu'\in\C}\rho(T_{\mu'}\backslash T_{k-1})\ge  \rho(T_{\mu}\backslash T_{k-1})\ge\delta>0,$$ which contradicts 
$m_k\rightarrow0$. Thus, we have shown that 
\begin{equation}\label{eq:mt}
\mu(T\cap T_{\mu})=1.
\end{equation}

Let us show that every $\mu\in\C$ is absolutely continuous with respect to $\nu$.
Indeed, fix any $\mu\in\C$ and suppose $\mu(A)>0$ for some $A\in\mathcal F$.
Then from~(\ref{eq:mt}) we have $\mu(A\cap T)>0$, and, by absolute continuity of $\mu$ with respect to $\rho$,
also $\rho(A\cap T)>0$. Since $T=\cup_{k\in\N}T_{k}$ we must have $\rho(A\cap T_k)>0$ for some $k\in\N$.
Since on the set $T_k$ the measure $\mu_k$ has non-zero density $f_{\mu_k}$ with respect to $\rho$, we must have $\mu_k(A\cap T_k)>0$.
(Indeed, $\mu_k(A\cap T_k)=\int_{A\cap T_k}f_{\mu_k}d\rho>0$.)
Hence,  
$$
\nu(A\cap T_k)\ge w_k \mu_k(A\cap T_k)> 0,
$$ so that $\nu(A)>0$.
Thus, $\mu$ is absolutely continuous with respect to $\nu$, and so $\nu$ predicts $\mu$ in total variation distance.
\end{proof}

\begin{theorem}\label{th:2} Let $\C$ be a set of probability measures on $\X^\infty$. If there is a measure $\rho$ such that $\rho$ predicts every $\mu\in\C$ in 
expected average KL divergence, then there is a sequence $\mu_k\in\C$, $k\in\N$ such that the measure $\nu:=\sum_{k\in\N} w_k\mu_k$ predicts every $\mu\in\C$ in 
expected average KL divergence, where $w_k$ are some positive weights.
\end{theorem}
A difference worth noting with respect to the formulation of Theorem~\ref{th:1} (apart from a different measure of divergence) is in that 
in the latter the weights $w_k$ can be chosen arbitrarily, while in Theorem~\ref{th:2} they can not.
In general, the statement ``$\sum_{k\in\N} w_k\nu_k$ predicts $\mu$ in expected average KL divergence for some choice of $w_k$, $k\in\N$''  does not imply
``$\sum_{k\in\N} w'_k\nu_k$ predicts $\mu$ in expected average KL divergence for every  summable sequence of positive $w_k', k\in\N$,'' while
the implication trivially holds true if the expected average KL divergence is replaced by the total variation. 
An interesting related question (which is beyond the scope of this paper) is how to chose the weights to optimize the behaviour of a predictor before asymptotic.

The idea of the proof is as follows. For every $\mu$ and every $n$ we consider the sets $T_\mu^n$ of those $x_{1..n}$ on which $\mu$ is greater 
than $\rho$. These sets have to have (from some $n$ on) a high  probability with respect to $\mu$. Then since $\rho$ predicts $\mu$ in 
expected average KL divergence, the $\rho$-probability of these sets cannot decrease exponentially fast (that is, it has to be quite large).
(The sequences  $\mu(x_{1..n})/\rho(x_{1..n})$, $n\in\N$ will play the role of densities of the proof of Theorem~\ref{th:1}, and the sets $T_\mu^n$
the role of sets $T_\mu$ on which the density is non-zero.)
We then use, for each given $n$ the same scheme to cover the set $\X^n$ with countably many $T_\mu^n$, as  was used in the proof of Theorem~\ref{th:1} 
to construct a countable covering of the set $\X^\infty$ , obtaining for each $n$ a predictor
$\nu_n$. Then the predictor $\nu$ is obtained as $\sum_{n\in\N} w_n \nu_n$, where the weights decrease subexponentially.
The latter fact ensures that, although the weights depend on $n$, they still play no role  asymptotically.
The technically most involved part of the proof is to show that the sets $T_\mu^n$ in asymptotic have sufficiently 
large weights in those
countable covers that we construct for each $n$. This is used to  demonstrate  the implication ``if a set has a high $\mu$ probability 
then its  $\rho$-probability does not decrease too fast, provided some regularity conditions.''
The proof is broken into the same steps as the (simpler) proof of Theorem~\ref{th:1}, to make the analogy explicit and the proof more understandable.
\begin{proof}
Define the weights $w_k:=wk^{-2}$, where $w$ is the normalizer $6/\pi^2$.

\noindent{\em Step 1: densities.} 
Define the sets 
\begin{equation}\label{eq:t}
T_\mu^n:=\left\{x_{1..n}\in \X^n: \mu(x_{1..n})\ge{1\over n}\rho(x_{1..n})\right\}.
\end{equation} 
Using Markov's inequality, we derive 
\begin{equation}\label{eq:mark}
\mu(\X^n\backslash T_\mu^n) 
 = \mu \left(\frac {\rho(x_{1..n})}{\mu(x_{1..n})} > n\right)\le {1\over n} E_\mu \frac {\rho(x_{1..n})}{\mu(x_{1..n})}={1\over n},
\end{equation}
so that $\mu(T_\mu^n)\to 1$.
(Note that  if $\mu$ is singular with respect to $\rho$, as is typically the case,  then $\frac{\rho(x_{1..n})}{\mu(x_{1..n})}$ converges to 0 $\mu$-a.e. and one
can replace ${1\over n}$ in~(\ref{eq:t}) by 1, while still having $\mu(T_\mu^n)\to1$.)

{\em Step 2n: a countable cover, time $n$.}
Fix an $n\in\N$. Define $m^n_1:=\max_{\mu\in\C}\rho(T_\mu^n)$ (since $\X^n$ are finite all suprema are reached). 
 Find any $\mu^n_1$ such that $\rho^n_1(T_{\mu^n_1}^n)=m^n_1$ and let
$T^n_1:=T^n_{\mu^n_1}$. For $k>1$, let $m^n_k:=\max_{\mu\in\C}\rho(T_\mu^n\backslash T^n_{k-1})$. If $m^n_k>0$, let $\mu^n_k$ be any $\mu\in\C$ such 
that $\rho(T_{\mu^n_k}^n\backslash T^n_{k-1})=m^n_k$, and let $T^n_k:=T^n_{k-1}\cup T^n_{\mu^n_k}$; otherwise let $T_k^n:=T_{k-1}^n$. Observe that 
(for each $n$) there is only a finite number of positive $m_k^n$,
since the set $\X^n$ is finite; let $K_n$ be the largest index $k$ such that $m_k^n>0$. Let 
\begin{equation}\label{eq:nun}
\nu_n:=\sum_{k=1}^{K_n} w_k\mu^n_k.
\end{equation}
As a result of this construction, for every $n\in\N$ every $k\le K_n$ and every  $x_{1..n}\in T^n_k$ 
using~(\ref{eq:t}) we obtain
\begin{equation}\label{eq:ext}
\nu_n(x_{1..n})\ge w_k{1\over n}\rho(x_{1..n}).
\end{equation}

{\em Step 2: the resulting predictor.}
Finally, define 
\begin{equation}\label{eq:nu}
\nu:={1\over 2}\gamma+{1\over2}\sum_{n\in\N}w_n\nu_n,
\end{equation}
 where $\gamma$ is the i.i.d. measure with equal probabilities of all $x\in\X$ 
(that is, $\gamma(x_{1..n})=|\X|^{-n}$ for every $n\in\N$ and every $x_{1..n}\in\X^n$). 
We will show that $\nu$  predicts every $\mu\in\C$, and 
then in the end of the proof (Step~r) we will show how to replace $\gamma$ by a combination of a countable set of elements of $\C$ (in fact, $\gamma$ 
is just a regularizer which ensures that $\nu$-probability of any word is never too close to~0). 

{\em Step 3: $\nu$ predicts every $\mu\in\C$.}
Fix any $\mu\in\C$. 
Introduce the parameters $\epsilon_\mu^n\in(0,1)$, $n\in\N$, to be defined later, and let $j_\mu^n:=1/\epsilon_\mu^n$.
Observe that $\rho(T^n_k\backslash T^n_{k-1})\ge \rho(T^n_{k+1}\backslash T^n_k)$, for any $k>1$ and any $n\in\N$, by definition of these sets.
Since the sets $T^n_k\backslash T^n_{k-1}$, $k\in\N$ are disjoint, we obtain $\rho(T^n_k\backslash T^n_{k-1})\le 1/k$. Hence,  $\rho(T_\mu^n\backslash T_j^n)\le \epsilon_\mu^n$ for some $j\le j_\mu^n$,
since  otherwise $m^n_j=\max_{\mu\in\C}\rho(T_\mu^n\backslash T^n_{j_\mu^n})> \epsilon_\mu^n$ so that  $\rho(T_{j_\mu^n+1}^n\backslash T^n_{j_\mu^n}) > \epsilon_\mu^n=1/j_\mu^n$, which is a contradiction. 
Thus,   
\begin{equation}\label{eq:tm}
\rho(T_\mu^n\backslash T_{j_\mu^n}^n)\le \epsilon_\mu^n.
\end{equation}
We can upper-bound $\mu(T_\mu^n\backslash T^n_{j^n_\mu})$ as follows. 
First, observe that
\begin{multline}\label{eq:mut}
d_n(\mu,\rho) 
=   -\sum_{x_{1..n}\in T^n_\mu\cap T^n_{j^n_\mu}}\mu(x_{1..n})\log\frac{\rho(x_{1..n})}{\mu(x_{1..n})} 
\\
-\sum_{x_{1..n}\in T^n_\mu\backslash  T^n_{j^n_\mu}}\mu(x_{1..n})\log\frac{\rho(x_{1..n})}{\mu(x_{1..n})} \\- \sum_{x_{1..n}\in \X^n\backslash T^n_\mu}\mu(x_{1..n})\log\frac{\rho(x_{1..n})}{\mu(x_{1..n})}
\\
=
I+II+III.
\end{multline}
Then, from~(\ref{eq:t}) we get 
\begin{equation}\label{eq:e1}
I\ge -\log n.
\end{equation}
Observe that for every $n\in\N$ and every set $A\subset \X^n$, using Jensen's inequality we can obtain
\begin{multline}\label{eq:jen}
-\sum_{x_{1..n}\in A}\mu(x_{1..n})\log\frac{\rho(x_{1..n})}{\mu(x_{1..n})}
=  -\mu(A)\sum_{x_{1..n}\in A}{1\over\mu(A)}\mu(x_{1..n})\log\frac{\rho(x_{1..n})}{\mu(x_{1..n})}
\\
\ge -\mu(A)\log{\rho(A)\over\mu(A)} \ge -\mu(A)\log\rho(A) -{1\over2}. 
\end{multline}
Thus, from~(\ref{eq:jen}) and~(\ref{eq:tm})
we get 
\begin{equation}\label{eq:e2}
II
\ge  -\mu(T_\mu^n\backslash T^n_{j^n_\mu}) \log\rho(T_\mu^n\backslash T^n_{j^n_\mu})- 1/2
\ge -\mu(T_\mu^n\backslash T^n_{j^n_\mu}) \log \epsilon_\mu^n - 1/2.
\end{equation}
Furthermore,  
\begin{multline}\label{eq:e3}
III\ge \sum_{x_{1..n}\in \X^n\backslash T^n_\mu}\mu(x_{1..n})\log\mu(x_{1..n}) 
\ge \mu(\X^n\backslash T^n_\mu)\log\frac{\mu(\X^n\backslash T^n_\mu)}{|\X^n\backslash T^n_\mu|}\\\ge -{1\over2} - \mu(\X^n\backslash T^n_\mu)n\log|\X|\ge-{1\over2} -\log|\X|,
\end{multline} 
where in the second inequality we have used the fact that entropy is maximized when all events are equiprobable, in the third one we used $|\X^n\backslash T^n_\mu|\le|\X|^n$,
while the last inequality follows from~(\ref{eq:mark}).
Combining~(\ref{eq:mut}) with the bounds~(\ref{eq:e1}), (\ref{eq:e2}) and~(\ref{eq:e3})  we obtain 
\begin{equation*}
d_n(\mu,\rho) \ge -\log n  -\mu(T_\mu^n\backslash T^n_{j^n_\mu}) \log \epsilon_\mu^n  - 1 - 
  \log|\X|,
\end{equation*}
so that
\begin{equation}\label{eq:mu2}
 \mu(T_\mu^n\backslash T^n_{j^n_\mu}) \le {1\over-\log \epsilon_\mu^n}\Big(d_n(\mu,\rho) +\log n  +1 + 
     \log|\X|\Big).
\end{equation}
Since $d_n(\mu,\rho)=o(n)$, 
 we can define the parameters $\epsilon^n_\mu$ in such a way that $-\log \epsilon^n_\mu=o(n)$ while
at the same time  the bound~(\ref{eq:mu2}) gives $\mu(T_\mu^n\backslash T^n_{j^n_\mu})=o(1)$. Fix such a choice of $\epsilon^n_\mu$.
Then,   using $\mu(T^n_\mu)\to1$,  we can conclude
\begin{equation}\label{eq:xt}
\mu(\X^n\backslash T^n_{j^n_\mu})\le \mu(\X^n\backslash T^n_{\mu})+ \mu(T^n_{\mu}\backslash T^n_{j^n_\mu}) =o(1).
\end{equation}

We proceed with the proof of $d_n(\mu,\nu)=o(n)$. 
For any $x_{1..n}\in T^n_{j_\mu^n}$  we have
\begin{equation}\label{eq:i}
\nu(x_{1..n})\ge {1\over 2}w_n\nu_n(x_{1..n})
\ge{1\over 2}w_n w_{j_{\mu}^n} {1\over n}\rho(x_{1..n})=\frac{w_nw}{2n}(\epsilon_\mu^n)^2\rho(x_{1..n}),
\end{equation}
where the first inequality follows from~(\ref{eq:nu}), the second from~(\ref{eq:ext}), and in the equality we have used $w_{j_{\mu}^n}=w/(j_{\mu}^n)^2$
and  $j_\mu^n=1/\epsilon^\mu_n$.
 Next we use the decomposition
\begin{multline}\label{eq:12}
d_n(\mu,\nu)= -\sum_{x_{1..n}\in T^n_{j_\mu^n}}\mu(x_{1..n})\log\frac{\nu(x_{1..n})}{\mu(x_{1..n})} \\ 
- \sum_{x_{1..n}\in \X^n\backslash T^n_{j_\mu^n}}\mu(x_{1..n})\log\frac{\nu(x_{1..n})}{\mu(x_{1..n})}  = I + II.
\end{multline}
From~(\ref{eq:i})  we find 
\begin{multline}\label{eq:1}
I\le -\log\left(\frac{w_nw}{2n}(\epsilon_\mu^n)^2\right)  - \sum_{x_{1..n}\in T^n_{j_\mu^n}}\mu(x_{1..n})\log\frac{\rho(x_{1..n})}{\mu(x_{1..n})}\\
=
(1+3\log n - 2\log\epsilon_\mu^n-2\log w) +\left(d_n(\mu,\rho)+ \sum_{x_{1..n}\in\X^n\backslash T^n_{j_\mu^n}}\mu(x_{1..n})\log\frac{\rho(x_{1..n})}{\mu(x_{1..n})}\right)
\\
\le o(n) -  \sum_{x_{1..n}\in\X^n\backslash T^n_{j_\mu^n}}\mu(x_{1..n})\log\mu(x_{1..n})\\ 
\le o(n)+\mu(\X^n\backslash T^n_{j_\mu^n})n\log|\X|=o(n),
\end{multline}
where in the second inequality we have used $-\log\epsilon_\mu^n=o(n)$ and $d_n(\mu,\rho)=o(n)$, in the last inequality we have again used the fact that the entropy is maximized when all events are equiprobable, 
while the last equality follows from~(\ref{eq:xt}). 
Moreover, from~(\ref{eq:nu}) we find
\begin{multline}\label{eq:2}
II\le \log 2 - \sum_{x_{1..n}\in\X^n\backslash  T^n_{j_\mu^n}}\mu(x_{1..n})\log\frac{\gamma(x_{1..n})}{\mu(x_{1..n})}
\\ 
\le 1 +n\mu(\X^n\backslash T^n_{j_\mu^n})\log|\X|=o(n),
\end{multline}
where in the last inequality we have used $\gamma(x_{1..n})=|\X|^{-n}$ and $\mu(x_{1..n})\le 1$, and the last equality follows from~(\ref{eq:xt}).

From~(\ref{eq:12}), (\ref{eq:1}) and~(\ref{eq:2}) we conclude ${1\over n}d_n(\nu,\mu)\to0$.

{\em Step r: the regularizer $\gamma$}. It remains to show that the  i.i.d. regularlizer $\gamma$ in the definition of $\nu$~(\ref{eq:nu}), can be replaced by a convex combination of a countably many elements from $\C$.
Indeed, for each $n\in\N$, denote
$$
A_n:=\{x_{1..n}\in \X^n: \exists\mu\in\C\ \mu(x_{1..n})\ne0\},
$$ and let $\mu_{x_{1..n}}:=\argmax_{\mu\in\C}\mu(x_{1..n})$ for each $x_{1..n}\in \X^n$.
Define 
$$
 \gamma_n'(x'_{1..n}):={1\over |A_n|}\sum_{x_{1..n}\in A_n}\mu_{x_{1..n}}(x'_{1..n}),
$$ for each
$x'_{1..n}\in A^n$, $n\in\N$, and let  $\gamma':=\sum_{k\in\N}w_k\gamma'_k$. 
For every $\mu\in\C$ we have 
$$
\gamma'(x_{1..n})\ge w_n|A_n|^{-1} \mu_{x_{1..n}}(x_{1..n})\ge w_n |\X|^{-n} \mu(x_{1..n})
$$ for
every $n\in\N$ and every $x_{1..n}\in A_n$, which clearly suffices to establish the bound $II=o(n)$ as in~(\ref{eq:2}).
\end{proof}
\section{Discussion}
For two measures of quality of prediction that we have considered, namely, total variation distance
and expected average KL divergence, we have shown that if a  prediction for a class $\C$ of measures
exists, then a predictor can also be obtained as a Bayesian mixture over a countable subset of $\C$.
The first  possible extension of these results that comes to mind is to find out 
whether the same  holds for other measures of performance, such as prediction in KL divergence
without time-averaging,  or with probability 1 rather then in expectation. 
Maybe the same 
results can be obtained in more general formulations, such as $f$-divergences of \cite{Csiszar:67}.

More generally, the questions we addressed in this work are a part of a larger problem:
given an arbitrary class $\C$ of stochastic processes, find the best predictor for it. 
One can approach this problem from other sides. For example, the first question one may wish to address is 
for which classes of processes a predictor
exists; see \cite{Ryabko08:pq+} for some sufficient conditions, such as separability of the class $\C$. Another approach is to 
identify the conditions which two measures $\mu$ and $\rho$ have to satisfy in order for $\rho$ to predict $\mu$.
For prediction in total variation such conditions have been identified \cite{Blackwell:62,Kalai:94} and, in particular, in the context of the present
 work, they turn out to be very useful. \cite{Kalai:94} also provides some characterization for the case of a weaker notion 
of prediction: difference between conditional probabilities of the next (several) outcomes (weak merging of opinions).
In \cite{Ryabko:08pqaml} some sufficient conditions are found for the case of prediction in expected average KL divergence,
and prediction in average KL divergence with probability 1. 
Of course, another  very natural approach to the general  problem posed above
is to try and find predictors (in the form of algorithms) for some particular classes of processes which 
are of practical interest.
Towards this end, the contribution of this work is in providing a specific form that some solution to this question has to have, 
if a solution exists: a Bayesian predictor whose prior is concentrated on a countable set. 
This is perhaps a rather simple form, which may be useful for constructing practical algorithms.

\bibliographystyle{apalike}

\end{document}